\documentclass[a4paper]{article}
\usepackage{iwslt18}
\usepackage{amssymb}
\usepackage{amsmath}
\usepackage{epsfig}

\usepackage{makecell}
\usepackage{multirow}
\usepackage{subfigure}
\usepackage{comment}

\usepackage{color}

\renewcommand{\emph}{\textit}

\def\reg{{\rm\ooalign{\hfil
     \raise.07ex\hbox{\scriptsize R}\hfl\crcr\mathhexbox20D}}}

\title{Multi-Source Neural Machine Translation with Data Augmentation}

\name{Yuta Nishimura\(^{1}\) , Katsuhito Sudoh\(^{1}\) , Graham Neubig\(^{2,1}\), Satoshi Nakamura\(^{1}\)}
\address{  \(^{1}\)Nara Institute of Science and Technology, 8916-5 Takayama-cho, Ikoma, Nara 630-0192, Japan\\
     \(^{2}\)Carnegie Mellon University, 5000 Forbes Avenue, Pittsburgh, PA 15213, USA\\
{\small  \tt \{nishimura.yuta.nn9, sudoh, s-nakamura\}@is.naist.jp} \\  
  \small \tt gneubig@cs.cmu.edu \\  
}

\begin{document}
\maketitle
\begin{abstract}
\emph{Multi-source} translation systems translate from multiple languages to a single target language.
By using information from these multiple sources, these systems achieve large gains in accuracy.
To train these systems, it is necessary to have corpora with parallel text in multiple sources and the target language. However, these corpora are rarely
complete in practice due to the difficulty of providing human translations
in \emph{all} of the relevant languages.
In this paper, we propose a data augmentation approach
to fill such incomplete parts using multi-source neural machine translation (NMT).
In our experiments, results varied over different language combinations
but significant gains were observed when using a source language similar to the target language.
\end{abstract}

\section{Introduction}

Machine Translation (MT) systems usually translate one source language to one target language.
However, in many real situations, there are multiple languages in the corpus of interest.
Examples of this situation include the multilingual official document collections
of the European parliament \cite{koehn2005epc} and the United Nations \cite{ZIEMSKI16.1195}.
These documents are manually translated into all official languages of the respective organizations.
Many methods have been proposed to use these multiple languages in translation systems to improve the translation accuracy \cite{dong-etc:2015:ACL,firat-cho-bengio:2016:N16-1,johonson-etc:2016:CoRR, ha2016toward}. 
In almost all cases, multilingual machine translation systems output better translations than one-to-one systems, as the system has access to multiple sources of information to reduce ambiguity in the target sentence structure or word choice.

However, in contrast to the more official document collections mentioned above where it is mandated that all translations in all languages, there are also more informal multilingual captions such as those of talks \cite{Cettolo-etc:2012:eamt} and movies \cite{Tiedemann:RANLP5}.
Because these are based on voluntary translation efforts,
large portions of them are not translated, especially into languages with a relatively small number of speakers.

\begin{figure}[t] 
    \centering
	\subfigure[Multi-source NMT with filling in a symbol \cite{Nishimura:2018:}]{
      \includegraphics[width=8.0cm]{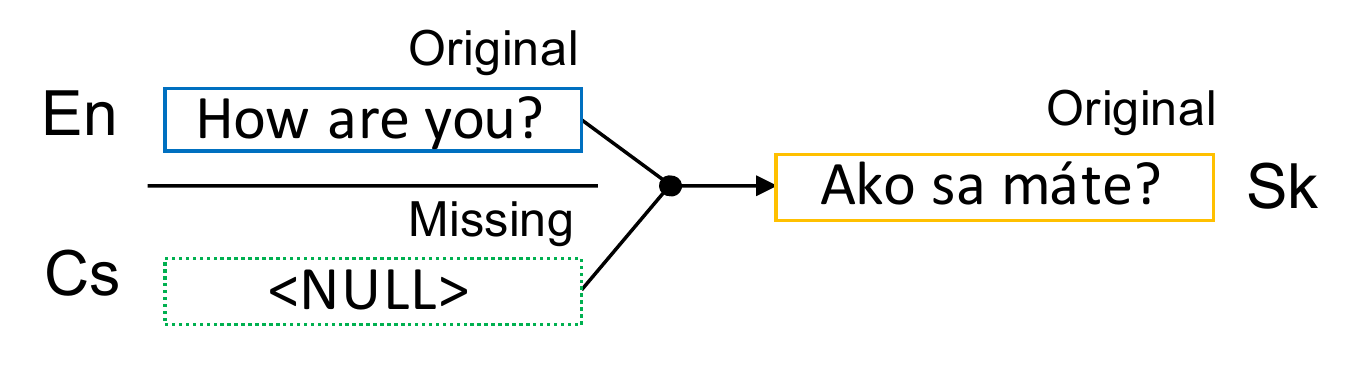}}
      \label{fig:multi-source_null}
	\subfigure[Proposed Method: Multi-source NMT with data augmentation]{
      \includegraphics[width=8.0cm]{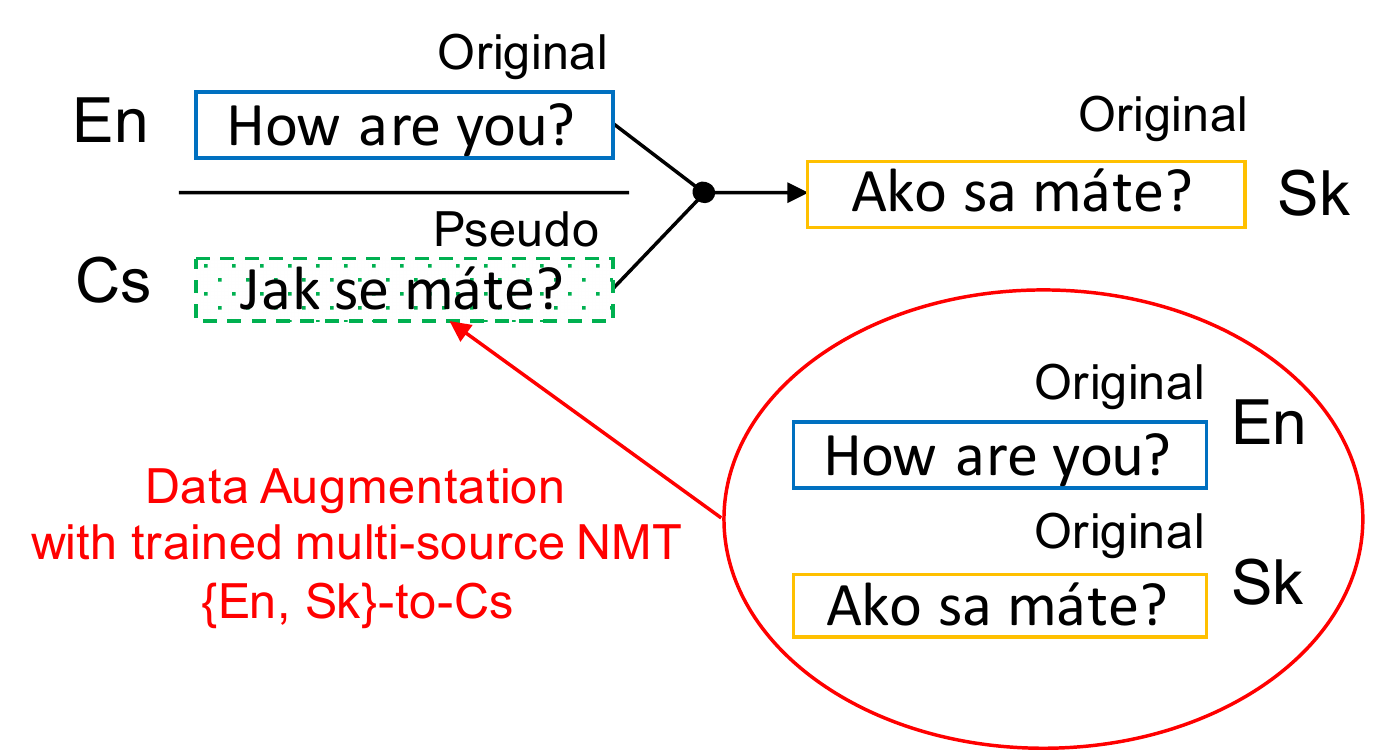}}
      \label{fig:multi-source_augment}
    \caption{Example of multi-source NMT with an incomplete corpus; The language pair is \{English, Czech\}-to-Slovak and the translation of Czech is missing.}
    \label{fig:multi-source_incomp}
\end{figure}

Nishimura \textit{et al.} \cite{Nishimura:2018:} have recently proposed a method for multi-source NMT that is able to deal with the case of missing source data encountered in these corpora. 
The implementation is simple: missing source translations are replaced with a special symbol $\langle$NULL$\rangle$ as shown in Figure~\ref{fig:multi-source_incomp}(a).
This method allows us to use incomplete corpora both at training time and test time, and multi-source NMT with this method was shown to achieve higher translation accuracy.
If the model is trained on corpora with a large number of $\langle$NULL$\rangle$ symbols on the source side, a large number of training examples will be different from test time, when we actually have multiple sources.
Thus, these examples will presumably be less useful in training a model intended to do multi-source translation.
In this paper, we propose an improved method for utilizing multi-source examples with missing data: using a pseudo-corpus whose missing translations are filled up with machine translation outputs using a trained multi-source NMT system as shown in Figure~\ref{fig:multi-source_incomp}(b).
Experimental results show that the proposed method is a more effective method to incorporate incomplete multilingual corpora, achieving improvements of up to about 2 BLEU over the previous method where each missing input sentence is replaced by $\langle$NULL$\rangle$.

\section{Related Work}
\subsection{Multi-source NMT}
There are two major approaches to multi-source NMT; multi-encoder NMT \cite{zoph-knight:2016:N16-1} and mixture of NMT Experts \cite{garmash-monz:2016:COLING}.
In this work, we focus on the multi-encoder NMT that showed better performance in most cases in Nishimura \textit{et al.} \cite{Nishimura:2018:}.

Multi-encoder NMT \cite{zoph-knight:2016:N16-1} is similar to the standard attentional NMT framework \cite{bahdanau-cho-bengio:2015:iclr} but uses multiple encoders corresponding to the source languages and a single decoder.

Suppose we have two LSTM-based encoders and
their hidden and cell states at the end of the inputs are $h_1$, $h_2$ and $c_1$, $c_2$, respectively.
Multi-encoder NMT initializes its decoder hidden state $h$ and cell state $c$ using these encoder states as follows:

\begin{equation}
h=\tanh (W_c[h_1;h_2])
\end{equation}
\begin{equation}
c = c_1 + c_2
\end{equation}

Attention is then defined over encoder states at each time step $t$
and resulting context vectors $d_t^1$ and $d_t^2$ are concatenated
together with the corresponding decoder hidden state $h_t$
to calculate the final context vector $\tilde{h_t}$. 

\begin{equation}
\tilde{h_t}=\tanh (W_c[h_t;d_t^1;d_t^2])
\end{equation}

Our multi-encoder NMT implementation is basically similar to the original one \cite{zoph-knight:2016:N16-1} but has a difference in its attention mechanism. We use global attention used in Nishimura \textit{et al.} \cite{Nishimura:2018:}, while Zoph and Knight used local-p attention. The global attention allows the decoder to look at everywhere in the input, while the local-p attention forces to focus on a part of the input \cite{luong-pham-manning:2015:EMNLP}.

\subsection{Data Augmentation for NMT}

Sennrich \textit{et al.} proposed a method to use monolingual training data in the target language
for training NMT systems, with no changes to the network architecture \cite{Sennrich-back:P16-1009}. 
It first trains a seed target-to-source NMT model using a parallel corpus
and then translates the monolingual target language sentences into the source language
to create a \textit{synthetic} parallel corpus.
It finally trains a source-to-target NMT model using the seed and synthetic parallel corpora.
This very simple method called \textit{back-translation} makes effective use of available resources, and achieves substantial gains.
Imamura \textit{et al.} proposed a method that enhances the encoder and attention using target monolingual corpora by generating mutliple source sentences via sampling as an extension of the back-translation \cite{W18-2707}.

There are also other approaches for data augmentation other than back-translation.
Wang \textit{et al.} proposed a method of randomly replacing words in both the source sentence and the target sentence with other random words from their corresponding vocabularies \cite{wang2018switchout}.
Kim and Rush proposed a sequence-level knowledge distillation in NMT that uses machine translation results by a large teacher model to train a small student model as well as ground-truth translations \cite{kim-rush:2016:EMNLP2016}.

Our work is an extension of the back-translation approach in multilingual situations
by generating pseudo-translations using multi-source NMT.

\section{Proposed Method}

We propose three types of data augmentation for multi-encoder NMT; ``fill-in'', ``fill-in and replace'' and ``fill-in and add.''
Firstly, we explain about the data requirements and overall framework using Figure~\ref{fig:multi-source_incomp}(b).
We used three languages; English, Czech and Slovak.
Our goal is to get the Slovak translation, and to do so we take three steps.
There are not any missing data in English translations, but Slovak and Czech translations have some missing data. 
In the first step, we train a multi-encoder NMT model (Source: English and Slovak, Target: Czech) to get Czech pseudo-translations using the baseline method, which is to replace a missing input sentence with a special symbol $\langle$NULL$\rangle$.
In the second step, we create Czech pseudo-translations using multi-encoder NMT which was trained on the first step.
We conducted three types of augmentation, which we introduce later. 
Finally in the third step, we switch the role of Czech and Slovak, in other words, we train a new multi-encoder NMT model (Source: English and Czech, Target: Slovak).
At this time, we use Slovak pseudo-translations in the source language side. 
This method is similar to back-translation but taking advantage of the fact that we have an additional source of knowledge (Czech or Slovak) when trying to augment the other language (Slovak or Czech respectively).

\begin{figure}[h!] 
    \centering
	\subfigure[fill-in]{
      \includegraphics[width=8.0cm]{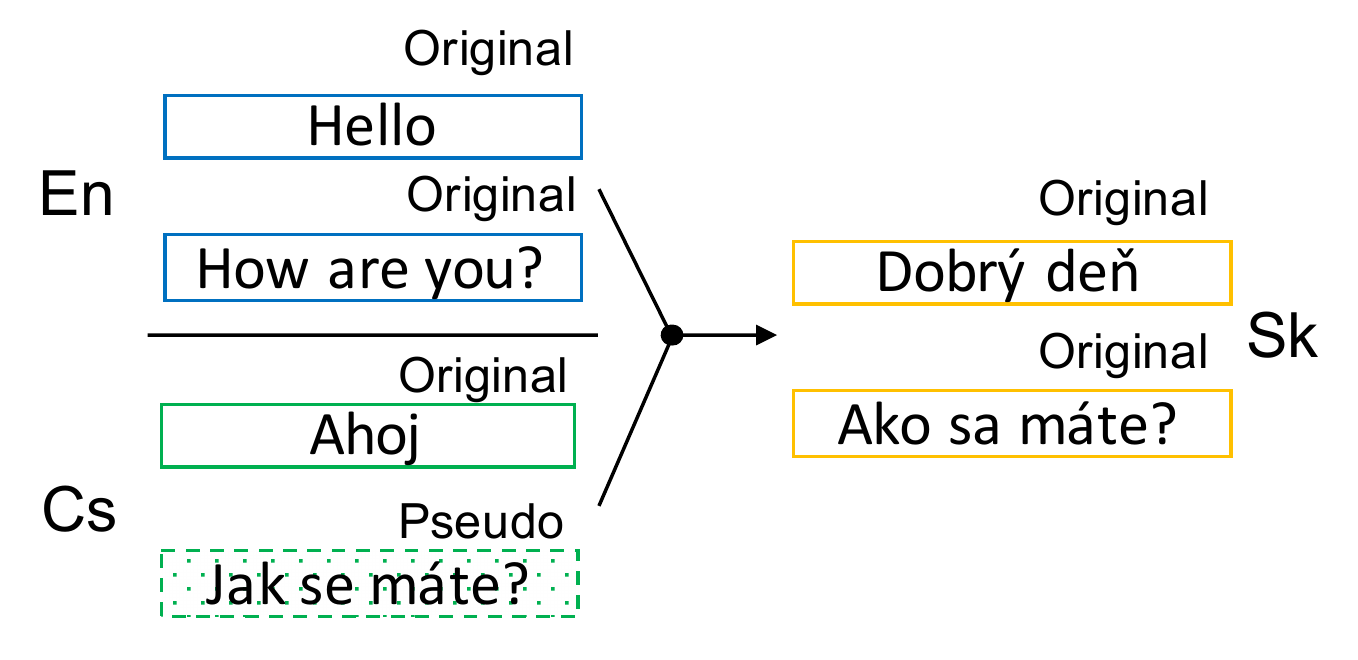}}
	\subfigure[fill-in and replace]{
      \includegraphics[width=8.0cm]{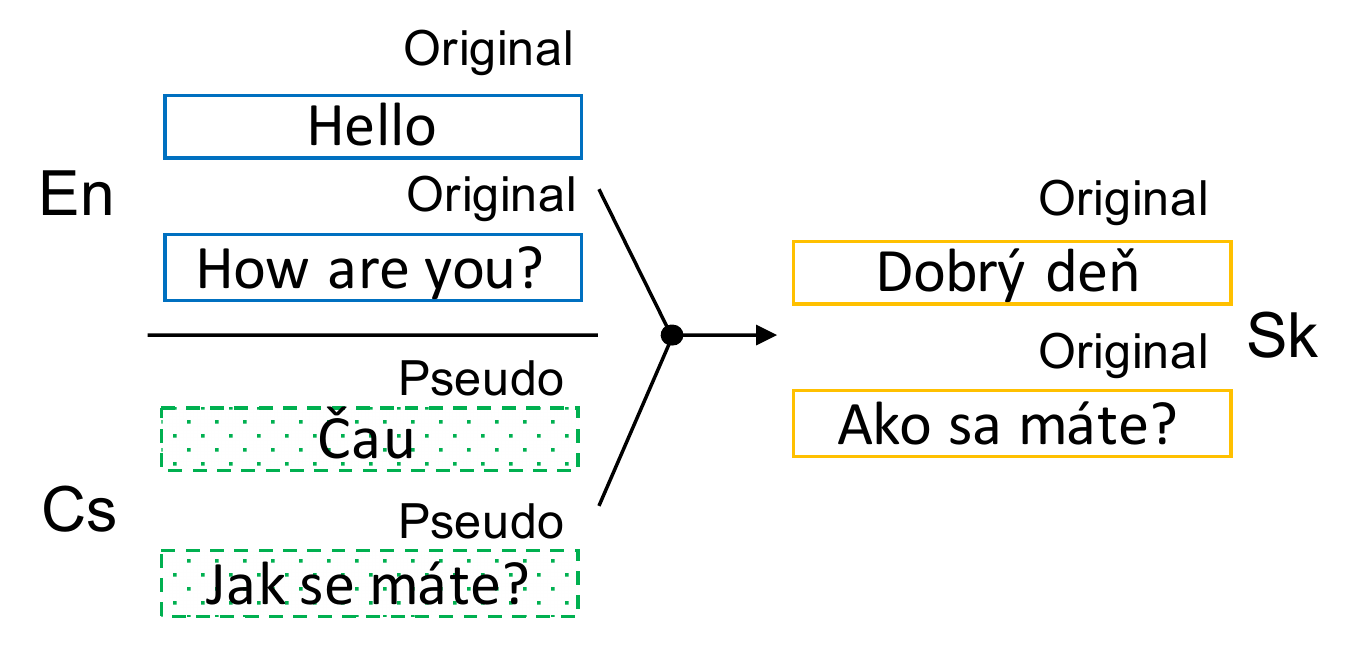}}
     \subfigure[fill-in and add]{
      \includegraphics[width=8.0cm]{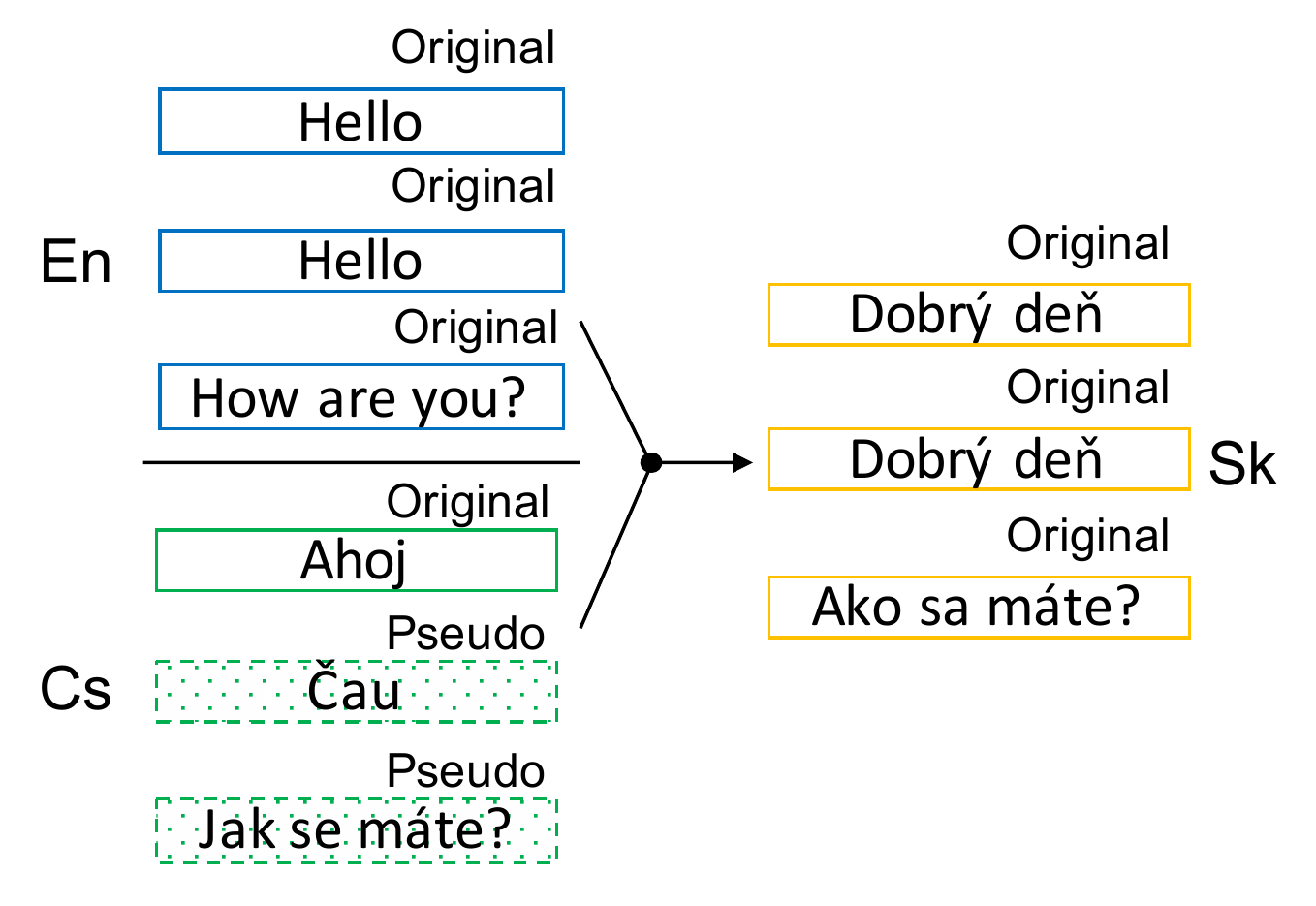}}
    \caption{Example of three types of augmentation; Language Pair is \{English, Czech\}-to-\{Slovak\} and Czech translation corresponding to ``How are you?'' is missisng. In this example, the dotted background indicates the pseudo-translation produced from multi-source NMT and the white background means the original translation.}
    \label{fig:augmentation_type}
\end{figure}

We next introduce three types of augmentation.
Figure~\ref{fig:augmentation_type} illustrates their examples in \{English, Czech\}-to-\{Slovak\} case where one Czech sentence is missing.

{
\setlength{\leftmargini}{0pt}
\begin{itemize}
\setlength{\parskip}{-3pt}
\setlength{\itemsep}{8pt}
\setlength{\itemindent}{0pt}

\item[] (a) \textbf{fill-in:} where only missing parts in the corpus are filled up with pseudo-translations. 
\item[] (b) \textbf{fill-in and replace:} where we both augment the missing part and replace original translations with pseudo-translations in the source language except English whose translations has not any missing data. %\gn{``except English'' is not clear. Is English L1? Isn't it the target, not the source?}. 
The motivation behind this method is not to use unreliable translation.
Morishita \textit{et al.} \cite{morishita2017ntt} demonstrated the effectiveness of applying back-translation for an unreliable part of a provided corpus.
Translations of TED talks are from many independent volunteers, so there may be some differences between translations other than original English, or even they may include some free or over-simplified translations.
We aim to fill such a gap using data augmentation.
  
\item[] (c) \textbf{fill-in and add:} where we both augment the missing part and added pseudo-translations from original translations in the source language except English. 
This helps prevent introduction of too much noise due to the complete replacement of original translations with pseudo-translations in the second method. 
\end{itemize}
}

\begin{comment}
\begin{figure*}[t] 
    \centering
	\subfigure[step 1: train the model \{L1,L2\}-to-L3 using the corpus which where missing translations are filled up with $\langle$NULL$\rangle$ (equal to baseline method Multi-encoder NMT with $\langle$NULL$\rangle$)]{
      \includegraphics[width=5.5cm]{./step_1.pdf}}
	\subfigure[step 2: Produce L3 pseudo-translations which is missing in the multilingual corpus]{
      \includegraphics[width=5.5cm]{./step_2.pdf}}
     \subfigure[step 3: Change source and target (L2 and L3) and train with the corpus which where missing translations are filled up with pseudo-translations]{
      \includegraphics[width=5.5cm]{./step_3.pdf}}
    \caption{Proposed Method; We suppose the corpus consisting of \{Talk A, Talk B, Talk C\}, and L1 has all of translations, but L2 doesn't have Talk A and L3 doesn't have Talk C.}
    \label{fig:proposed method}
\end{figure*}
\end{comment}

\section{Experiment}
We conducted MT experiments to examine the performance of the proposed method
using actual multilingual corpora of TED Talks.

\subsection{Data}
We used a collection of transcriptions of TED Talks and their multilingual translations.
The numbers of these voluntary translations differs significantly by language.
We chose three different language sets for the experiments:
\{English (en), Croatian (hr), Serbian (sr)\},
\{English (en), Slovak (sk), Czech (cs)\},
and \{English (en), Vietnamese (vi), Indonesian (id)\}.
Since the great majority of TED talks are in English,
the experiments were designed for the translation from English to another language
with the help of the other language in the language set, with no missing portions in the English sentences.
Table~\ref{table:data_detail} shows the number of training sentences for each language set.
At test time, we experiment with a complete corpus with both source sentences represented, as this is the sort of multi-source translation setting that we are aiming to create models for.

\begin{table}
\caption{``train" shows the number of available training sentences, and ``missing" shows the number and the fraction of missing sentences in comparison with English ones.}
\vspace{2mm}
\centering{
\begin{tabular}{|c|c|c|c|} \hline
Pair & Trg & train & missing \\ \hline
\multirow{2}{*}{en-hr/sr} & hr & 118949 & 35564 (29.9$\%$)  \\ \cline{2-4}
& sr & 133558 & 50203 (37.6$\%$)  \\ \hline
\multirow{2}{*}{en-sk/cs} & sk & 100600 & 58602 (57.7$\%$) \\ \cline{2-4}
& cs & 59918 &  17380 (29.0$\%$) \\ \hline
\multirow{2}{*}{en-vi/id} & vi & 160984 & 87816 (54.5$\%$)\\ \cline{2-4}
& id & 82592 & 9424 (11.4$\%$) \\ \hline
\end{tabular}}
\label{table:data_detail}
\end{table}

\subsection{Baseline Methods}
We compared the proposed methods with the following three baseline methods.
{
\setlength{\leftmargini}{0pt}
\begin{itemize}
\setlength{\parskip}{-3pt}
\setlength{\itemsep}{8pt}
\setlength{\itemindent}{0pt}

\item[] \textbf{One-to-one NMT:} a standard NMT model from one source language to another target language.
The source language is fixed to English in the experiments.
If the target language part is missing in the parallel corpus, such sentences pairs cannot be used in training so they are excluded from the training set.
\item[] \textbf{Multi-encoder NMT with back-translation:} a multi-encoder NMT system using English-to-X NMT to fill up the missing parts in the other source language X.\footnote{This is not exactly \textit{back}-translation because the pseudo-translations are not from the target language but from the other source language (English) in our multi-source condition. But we use this familiar term here for simplicity.} 
\item[] \textbf{Multi-encoder NMT with $\langle$NULL$\rangle$:} a multi-encoder NMT system using a special symbol $\langle$NULL$\rangle$ to fill up the missing parts in the other source language X \cite{Nishimura:2018:}.
\end{itemize}
}

\subsection{NMT settings} 
NMT \footnote{We used primitiv as a neural network toolkit. https://github.com/primitiv/primitiv} settings are the same for all the methods in the experiments.
We use bidirectional LSTM encoders \cite{bahdanau-cho-bengio:2015:iclr}, and global attention and input feeding for the NMT model \cite{luong-pham-manning:2015:EMNLP}.
The number of dimensions is set to 512 for the hidden and embedding layers.
Subword segmentation was applied using SentencePiece \cite{Kudo:P18-1007}.
We trained one subword segmentation model for English and another shared between the other two languages in the language set
because the amount of training data for the languages other than English was small.
For parameter optimization, we used Adam \cite{kingma-ba:2015:iclr} with gradient clipping of 5.
We performed early stopping, saving parameter values that had the best log likelihoods on the validation data and used them when decoding test data.

\subsection{Results}
Table~\ref{table:main_result} shows the results in BLEU \cite{papineni-etc:2002:ACL}. 
We can see that our proposed methods demonstrate larger gains in BLEU than baseline methods in two language sets: \{English, Croatian, Serbian\}, \{English, Slovak, Czech\}. 
On these pairs, we can say that our proposed method is an effective way for using incomplete multilingual corpora, exceeding other reasonably strong baselines.
However, in \{English, Vietnamese, Indonesian\}, our proposed methods obtained lower scores than the baseline methods.
We observed that the improvement by the use of multi-encoder NMT against one-to-one NMT in the baseline was significantly smaller than the other language sets, so multi-encoder NMT was not as effective compared to one-to-one NMT in the first place. 
Our proposed method is affected by which languages to use, and the proposed method is likely more effective for similar language pairs because the expected accuracy of the pseudo-translation gets better by the help of lexical and syntactic similarity including shared subword entries.

\begin{table*}
\caption{Main results in BLEU for English-Croatian/Serbian (en-hr/sr), English-Slovak/Czech (en-sk/cs), and English-Vietnamese/Indonesian (en-vi/id). 
}
\vspace{2mm}
\centering{
\begin{tabular}{|c|c|c|c|c|c|c|c|} \hline
 &  & \multicolumn{3}{|c|}{baseline method} & \multicolumn{3}{|c|}{proposed method} \\ \cline{3-8}
Pair & Trg & \thead{one-to-one \\ (En-to-Trg)} & \thead{multi-encoder NMT \\ (fill up with symbol)} & \thead{multi-encoder NMT \\ (back translation)} & \thead{fill-in} & \thead{fill-in and \\ replace} & \thead{fill-in \\ and  add} \\ \hline \hline
\multirow{2}{*}{en-hr/sr} & hr & 20.21 & 28.18 & 27.57  & 29.17 & 29.37 & \textbf{29.40} \\ \cline{2-8}
& sr & 16.42 & 23.85 & 22.73 & 24.41 & \textbf{24.96} & 24.15 \\ \hline \hline
\multirow{2}{*}{en-sk/cs} & sk & 13.79 & 20.27 & 19.83 & 20.26 & 20.43 & \textbf{20.59}  \\ \cline{2-8}
& cs & 14.72 & 19.88 & 19.54 & 20.78  & \textbf{20.90} & 20.61  \\ \hline \hline
\multirow{2}{*}{en-vi/id} & vi & 24.60 & 25.70 & 26.66 & \textbf{26.73} & 26.48 & 26.32 \\ \cline{2-8}
& id & 24.89 & \textbf{26.89} & 26.34 & 26.40  & 25.73 &  26.21  \\ \hline
\end{tabular}}
\label{table:main_result}
\end{table*}

\section{Discussion}
\subsection{Different Types of Augmentation}
We examined three types of augmentation: ``fill-in'', ``fill-in and replace'', ``fill-in and add''. 
In Table~\ref{table:main_result}, we can see that there were no significant differences among them, despite the fact that their training data were very different from each other.
We conducted additional experiments using incomplete corpora with lower quality augmentation by one-to-one NMT to investigate the differences of the three types of augmentation.
We created three types of pseudo-multilingual corpora using back-translation from one-to-one NMT
and trained multi-encoder NMT models using them.
Our expectation here was that the aggressive use of low quality pseudo-translations caused to contaminate the training data and to decrease the translation accuracy.

Table~\ref{table:augmentation_types} shows the results.
In \{English, Croatian, Serbian\} and \{English, Slovak, Czech\},
we obtained significant drop in BLEU scores with the aggressive strategies (``fill-in and replace'' and ``fill-in and add''),
while there are few differences in \{English, Vietnamese, Indonesian\}.
One possible reason is that the quality of pseudo-translations by one-to-one NMT in Indonesian and Vietnamese was better than the other languages;
in other words, the BLEU from one-to-one NMT in Table~\ref{table:main_result} was sufficiently good without multi-source NMT.
Thus the translation performance for Croatian, Serbian, Slovak and Czech could not improve in the experiments here due to \textit{noisy} pseudo-translations of those languages.
Contrary, the BLEU from ``fill-in and add'' was the highest when the target language was Indonesian.
We hypothesize that this is due to much smaller fraction of the missing parts in Indonesian corpus as shown in Table~\ref{table:data_detail}, so there should be little room for improvement if we fill in only the missing parts even if the accuracy of the pseudo-translations is relatively high.

\begin{table}
\caption{The difference of three types of augmentation in BLEU for English-Croatian/Serbian (en-hr/sr), English-Slovak/Czech (en-sk/cs), and English-Vietnamese/Indonesian (en-vi/id). We used one-to-one model to produce pseudo-translations.}
\vspace{2mm}
\centering{
\begin{tabular}{|c|c|c|c|c|} \hline
 &  & \multicolumn{3}{|c|}{multi-encoder NMT (back-translation)} \\ \cline{3-5}
Pair & Trg & \thead{fill-in} & \thead{fill-in and \\ replace} & \thead{fill-in \\ and add} \\ \hline \hline
\multirow{2}{*}{en-hr/sr} & hr & \textbf{27.57} & 24.05 & 24.79  \\ \cline{2-5}
& sr & \textbf{22.73} & 17.77 & 22.02  \\ \hline \hline
\multirow{2}{*}{en-sk/cs} & sk & \textbf{19.83} & 16.75 & 18.16   \\ \cline{2-5}
& cs & \textbf{19.54} & 17.04 & 18.40  \\ \hline \hline
\multirow{2}{*}{en-vi/id} & vi & \textbf{26.66} & 26.39 & 26.65  \\ \cline{2-5}
& id & 26.34 & 23.90 & \textbf{26.67}  \\ \hline
\end{tabular}}
\label{table:augmentation_types}
\end{table}

\subsection{Iterative Augmentation}
It can be noted that if we have a better multi-source NMT system, it can be used to produce better pseudo-translations.
This leads to a natural iterative training procedure where we alternatively update the multi-source NMT systems into the two target languages.

Table~\ref{table:iterate} shows the results of \{English, Croatian, Serbian\}.
We found that this produced negative results; BLEU decreased gradually in every step.
We observed very similar results in the other language pairs, while we omit the actual numbers here.
This indicates that the iterative training may be introducing more noise than it is yielding improvements, and thus may be less promising than initially hypothesized.

\begin{table*}[h!]
\caption{BLEU (and BLEU gains compared to step 1) in each step of iterative augmentation.}
\vspace{2mm}
\centering{
\begin{tabular}{|c|c|c|c|c|c|} \hline
Pair & Trg & step 1 & step 2 & step 3 & step 4 \\ \hline \hline
\multirow{2}{*}{en-hr/sr} & hr & 29.17 (+0.00) & 29.03 (-0.14)  & 29.10 (-0.07) & 29.05 (-0.12) \\ \cline{2-6}
& sr & 24.41 (+0.00) & 24.18 (-0.23) & 24.17 (-0.24) & 23.95 (-0.46)  \\ \hline
\end{tabular}}
\label{table:iterate}
\end{table*}

\begin{comment}
\begin{figure}[t]
	\includegraphics[width=7.5cm]{./iterate.png}
    \caption{BLEU in each step of augmentation}
    \label{fig:iterate}
\end{figure}
\end{comment}

\begin{comment}
\subsection{Test with incomplete corpus}
[In table 2, I conducted test with complete corpus, so I will write that the result with incomplete corpus]
[Table 4 shows the result. But the result is not good, so I think we don't have to write this discussion]

\begin{table*}
\caption{Test with incomplete corpus in BLEU for English-Croatian/Serbian (en-hr/sr), English-Slovak/Czech (en-sk/cs), and English-Vietnamese/Indonesian (en-vi/id). 
}
\vspace{2mm}
\centering{
\begin{tabular}{|c|c|c|c|c|c|c|c|} \hline
 &  &  & \multicolumn{2}{|c|}{fill up with NULL} & \multicolumn{2}{|c|}{back-translation} \\ \cline{4-7}
Pair & Trg & \thead{one-to-one \\ (En-to-Trg)} & \thead{multi-encoder NMT \\ (baseline)} & \thead{multi-encoder NMT \\ (proposed method)} & \thead{multi-encoder NMT \\ (baseline)} & \thead{multi-encoder NMT \\ (proposed method) } \\ \hline\hline
\multirow{2}{*}{en-hr/sr} & hr & 21.99 & 22.12  & 18.40 & 20.57 & 20.20 \\ \cline{2-7}
& sr & 14.51 & 14.09 & 2.50 & 13.96 & 14.38 \\ \hline \hline
\multirow{2}{*}{en-sk/cs} & sk & 13.14 & & & &   \\ \cline{2-7}
& cs & 13.92 & & & &    \\ \hline \hline
\multirow{2}{*}{en-vi/id} & vi & 22.80 & & & &  \\ \cline{2-7}
& id & 26.80 & & & &   \\ \hline
\end{tabular}}
\label{table:incomp_test}
\end{table*}
\end{comment}

\subsection{Non-parallelism}
A problem in the use of multilingual corpora is non-parallelism.
In case of TED multilingual captions,
they are translated from English transcripts independently by many volunteers,
which may cause some differences in details of the translation in the various target languages.
For example in \{English, Croatian, Serbian\},
Croatian and Serbian translations may not be completely parallel.
Table~\ref{table:examples} shows such an example
where the Serbian translation does not have a phrase corresponding to ``let me.''
This kind of non-parallelism may be resolved by overriding such translations with pseudo-translations with ``fill-in and replace'' and ``fill-in and add''.
Here, the Serbian pseudo-translation includes the corresponding phrase ``Dozvolite mi'' and can be used to compensate for the missing information.
This would be one possible reason of the improvements by ``fill-up and replace'' or ``fill-up and add''.

 \begin{table*}[t]
   \caption{Example of the Serbian pseudo-translation. This pseudo-translation is the output of \{English, Croatian\}-to-Serbian translation.}
  \begin{center}
   \begin{tabular}{|l||l|}
    \hline
    Type             & Sentence \\ \hline\hline

    Original (En)      & So \textbf{let me} conclude with just a remark to bring it back to the theme of choices. \\
    Original (Sr)      & Da zaključim jednom konstatacijom kojom se vraćam na temu izbora. \\
    Pseudo (Sr)        & \textbf{Dozvolite mi} da zaključim samo jednom opaskom, da se vratim na temu izbora. \\ \hline
   \end{tabular}
   \label{table:examples}
  \end{center}
 \end{table*}

\section{Conclusions}
In this paper, we examined data augmentation of incomplete multilingual corpora in multi-source NMT.
We proposed three types of augmentation; fill-in, fill-in and replace, fill-in and add.
Our proposed methods proved better than baseline system using the corpus where missing part was filled up with ``$\langle$NULL$\rangle$'', although  results depended on the language pair.
One limitation in the current experiments with a set of three languages was that missing parts in the test sets could not be filled in.
This can be resolved if we use more languages, and we will investigate this in future work.

\section{Acknowledgements}
Part of this work was supported by JSPS KAKENHI Grant Numbers and JP16H05873
and JP17H06101.

\bibliographystyle{IEEEtran}
\bibliography{iwslt2018}

\end{document}